\documentclass[journal]{journal}
\usepackage{ifpdf}

\usepackage{natbib}
\setcitestyle{numbers,open={[},close={]}}

\ifCLASSINFOpdf
\usepackage[pdftex]{graphicx}
\else
\fi
\usepackage[cmex10]{amsmath}
\usepackage{algorithmic}

\usepackage{array}

\usepackage{mdwmath}
\usepackage{mdwtab}

\usepackage{eqparbox}

\usepackage[utf8]{inputenc} 
\usepackage[T1]{fontenc}    
\usepackage{hyperref}       
\usepackage{url}            
\usepackage{booktabs}       
\usepackage{amsfonts}       
\usepackage{nicefrac}       
\usepackage{microtype}      
\usepackage{xcolor}         
\usepackage{url}            
\usepackage{booktabs}       
\usepackage{amsfonts}       
\usepackage{nicefrac}       
\usepackage{microtype}      
\usepackage{listings}       
\usepackage{xcolor}
\usepackage{caption}
\usepackage{subcaption}
\usepackage{multicol}
\usepackage{siunitx}
\usepackage{booktabs}
\usepackage{array}
\newcolumntype{C}[1]{>{\centering\arraybackslash}p{#1}}
\definecolor{codegreen}{rgb}{0,0.6,0}
\definecolor{codegray}{rgb}{0.5,0.5,0.5}
\definecolor{codepurple}{rgb}{0.58,0,0.82}
\definecolor{backcolour}{rgb}{0.95,0.95,0.92}
\lstdefinestyle{mystyle}{
    backgroundcolor=\color{backcolour},   
    commentstyle=\color{codegreen},
    keywordstyle=\color{magenta},
    numberstyle=\tiny\color{codegray},
    stringstyle=\color{codepurple},
    basicstyle=\ttfamily\tiny,
    breakatwhitespace=false,         
    breaklines=true,                 
    captionpos=b,                    
    keepspaces=true,                 
    numbers=left,                    
    numbersep=5pt,                  
    showspaces=false,                
    showstringspaces=false,
    showtabs=false,                  
    tabsize=2}
\usepackage[export]{adjustbox} 
\usepackage{graphicx}

\usepackage{caption}
\usepackage{stfloats}
\usepackage{url}

\usepackage[utf8]{inputenc} 
\usepackage[T1]{fontenc}    
\usepackage{hyperref}       
\usepackage{url}            
\usepackage{booktabs}       
\usepackage{amsfonts}       
\usepackage{nicefrac}       
\usepackage{microtype}      
\usepackage{xcolor}         
\usepackage{lipsum}
\usepackage{graphicx}
\usepackage[export]{adjustbox}

\begin{document}
%
\title{Integrating Distributed Architectures in Highly Modular RL Libraries}
%
%
%
\author{Albert Bou\textsuperscript{1}, Sebastian Dittert\textsuperscript{1}, and Gianni de Fabritiis\textsuperscript{1,2} \\
\thanks{Computational Science Laboratory, Universitat Pompeu Fabra (UPF) \textsuperscript{1}} \thanks{ICREA \textsuperscript{2}}}

\maketitle
\thispagestyle{empty}

\begin{abstract}
  Advancing reinforcement learning (RL) requires tools that are flexible enough to easily prototype new methods while avoiding impractically slow experimental turnaround times. To match the first requirement, the most popular RL libraries advocate for highly modular agent composability, which facilitates experimentation and development. To solve challenging environments within reasonable time frames, scaling RL to large sampling and computing resources has proved a successful strategy.
  However, this capability has been so far difficult to combine with modularity. In this work, we explore design choices to allow agent composability both at a local and distributed level of execution. We propose a versatile approach that allows the definition of RL agents at different scales through independent reusable components. We demonstrate experimentally that our design choices allow us to reproduce classical benchmarks, explore multiple distributed architectures, and solve novel and complex environments while giving full control to the user in the agent definition and training scheme definition. We believe this work can provide useful insights to the next generation of RL libraries.
\end{abstract}


\bibliographystyle{unsrtnat} 

%
\IEEEpeerreviewmaketitle

The domain of reinforcement learning (RL) has experienced significant advancements in recent years, transitioning from game-based applications \cite{mnih2015human},\cite{ OpenAI_dota}, \cite{silver2018general}, \cite{vinyals2019grandmaster} to simulators \cite{coumans2017pybullet}, \cite{crosby2019animal}, \cite{juliani2019obstacle}, \cite{guss2019minerl}, \cite{savva2019habitat} and to addressing practical challenges encountered in real-world scenarios such as drug discovery \cite{blaschke2020reinvent}, controlling cooling systems \cite{luo2022controlling} or chip design \cite{mirhoseini2021graph}. This progress has propelled RL to emerge as an increasingly influential discipline. However, the acceptance and widespread utilization of general RL libraries has been slow and fragmented. One contributing factor to this phenomenon is the intricate nature of modern RL algorithms, which typically consist of numerous interdependent and heterogeneous algorithmic components. These components include data buffers, neural networks, training loss functions, and more. The complex interplay between these components often poses challenges to developing general well-integrated RL libraries, leading many researchers to resort to writing custom code tailored to their specific applications.

To tackle this challenge, prominent libraries in the field such as Tianshou \cite{tianshou} or RLlib \cite{liang2017rllib} have embraced highly modular methodologies, enabling component reusability and limiting code complexity. This approach provides a practical solution, especially in research settings, by allowing processes to be segregated into independent components. These components can then be combined in a flexible manner to construct RL agents. Libraries such as TorchRL \cite{TorchRL} have further advanced this concept by providing completely standalone components and introducing robust data carriers to improve class-to-class communication. These features facilitate seamless interaction between individual components. As a result, practitioners can leverage tested and reusable modules and combine them with novel ideas implemented as new components, offering a balance between leveraging existing knowledge and exploring innovative approaches.
 
Nevertheless, an additional still unsolved challenge stems from the intrinsic sample inefficiency of most RL algorithms, which often demand a substantial volume of samples to tackle intricate problems effectively within reasonable time frames. In recent years, researchers have turned to scale-up algorithms by distributing data collection and other computationally demanding tasks, an approach that has shown promise in works like IMPALA \cite{espeholt2018impala}, R2D2 \cite{kapturowski2018recurrent}, DPPO \cite{heess2017emergence}, DDPPO \cite{wijmans2019dd} or RAPID \cite{OpenAI_dota}. These results have consequently made scalability a desirable feature in RL libraries. However, integrating distributed training from a modular perspective has proven to be challenging thus far. Currently, most libraries tend to adopt a rigid approach to scalability, even if the rest of the library is highly modular. They rely on high-level methods that handle the distributed intricacies internally, which can limit users' ability to exercise fine-grained control over this relevant training aspect. However, while most distributed works are supported by different underlying distributed schemes, they can be broken down into common reusable components, allowing for the adoption of a modular approach.

In this study, we focus on finding good design choices to integrate distributed training into a modular framework. 
To achieve this, we start by implementing PyTorch-based \cite{paszke2019pytorch} agents from the seamless combination of individual agent components, such as storages, losses, or actors. This combination empowers us to create a diverse set of RL methods, including on-policy, off-policy, and model-based approaches. Components can be reused in different agents and new components can be easily designed or extended and combined with those already available without any need to change the library's internal code. This is akin to the most successful RL libraries. Additionally, we develop scheme components, which encompass various functionalities and distributed options. By combining these components, a diverse array of training schemes can be defined with the independence of the agent components, extending the inherent benefits of composable agents to the context of distributed training. We validate our design choices by testing a variety of training schemes against classic benchmarks. We also test their capacity to accelerate training processes in increasingly large clusters and obtain the highest to-date test performance on the Obstacle Tower Unity3D challenge environment (\cite{juliani2019obstacle}). In doing so, our goal is to provide valuable insights into successfully integrating agent and scheme composability and to offer guidance for future projects on making informed design choices.

Our work contributes in several key ways. Firstly, we make an analysis of distributed RL requirements and propose a set of components that allow for the creation of distributed training schemes at various levels of execution following a modular paradigm. Secondly, we investigate and test our design choices, showcasing its ability to accurately replicate classical benchmarks and to effectively address novel problems while granting complete control over the scaling option and being transparent to the user. 
Lastly, our work demonstrates the advantages of integrating distributed applications using the paradigm of independent, reusable, and interchangeable components, such as code comprehensibility and simplification. We firmly believe our work can offer valuable insights for the future generations of RL libraries, aimed at achieving widespread adoption among the research and development community.

\section{Related Work}

The advancements in RL have led to the appearance of a wide array of techniques. To enhance accessibility and comparability of RL methods, numerous libraries have been introduced with the objective to address the complexities associated with the implementation, training, and assessment of both existing and emergent RL algorithms \cite{stable-baselines3}, \cite{ding}, \cite{huang2022cleanrl}, \cite{garage}, \cite{LanctotEtAl2019OpenSpiel}, \cite{castro2018dopamine}, \cite{TFAgents}. 
However, despite the abundance of RL libraries, practitioners still face difficulties in finding suitable options that allow for the exploration of new methods while facilitating flexible scalability in training.

For example, many formerly prevalent libraries have gradually fallen into obsolescence, with several no longer receiving regular maintenance \cite{tensorforce},\cite{ plappert2016kerasrl} ,\cite{duan2016benchmarking}, \cite{caspi_itai_2017_1134899}. 
Other libraries fall short in terms of modularity, presenting hurdles for researchers and developers intending to integrate custom algorithmic components for application and method testing \cite{huang2022cleanrl}, \cite{castro2018dopamine}.
Furthermore, while certain libraries adopt modularity as a design principle, they lack considerations for distributed training, and focus predominantly on single-threaded implementations \cite{JMLR:v22:18-056}, \cite{baselines}, \cite{tensorforce}, \cite{tianshou}, \cite{garage}. 

\begin{figure*}[!ht]
\minipage{0.33\textwidth}
  \includegraphics[width=\linewidth, frame]{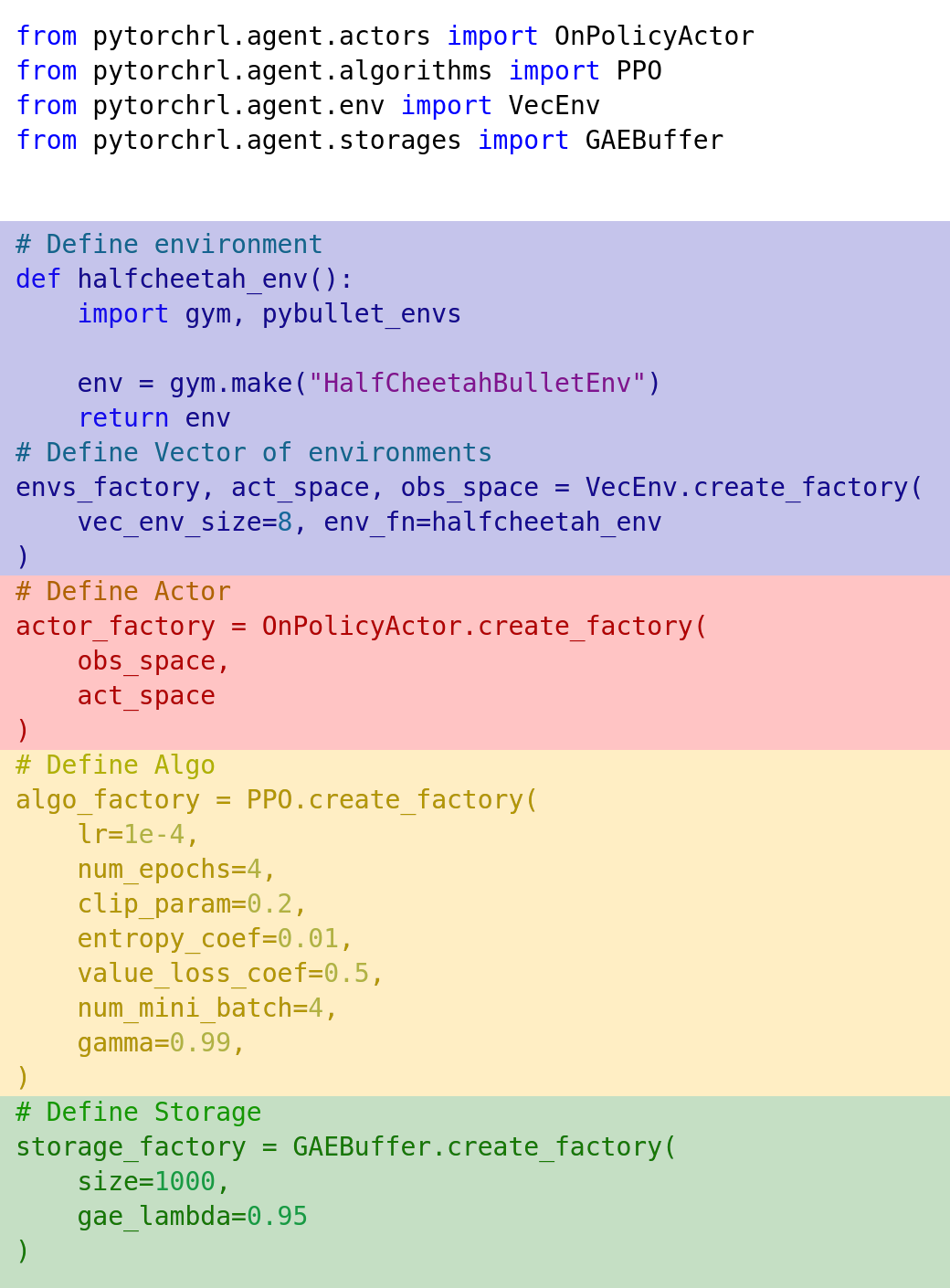}
  \caption{PPO example}\label{fig:ppo_example}
\endminipage\hfill
\minipage{0.33\textwidth}
  \includegraphics[width=\linewidth, frame]{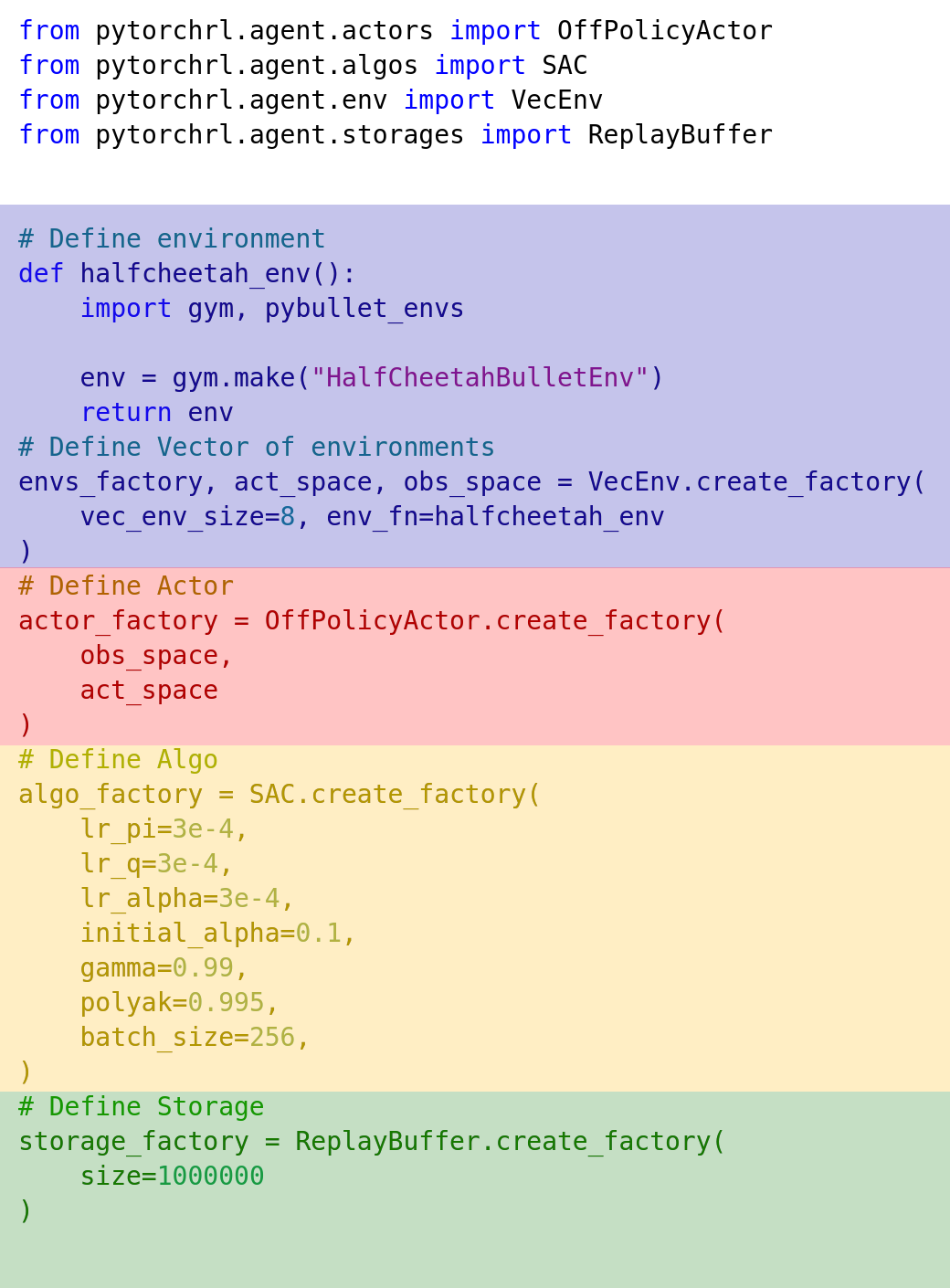}
  \caption{SAC example}\label{fig:sac_example}
\endminipage\hfill
\minipage{0.33\textwidth}%
  \includegraphics[width=\linewidth,frame]{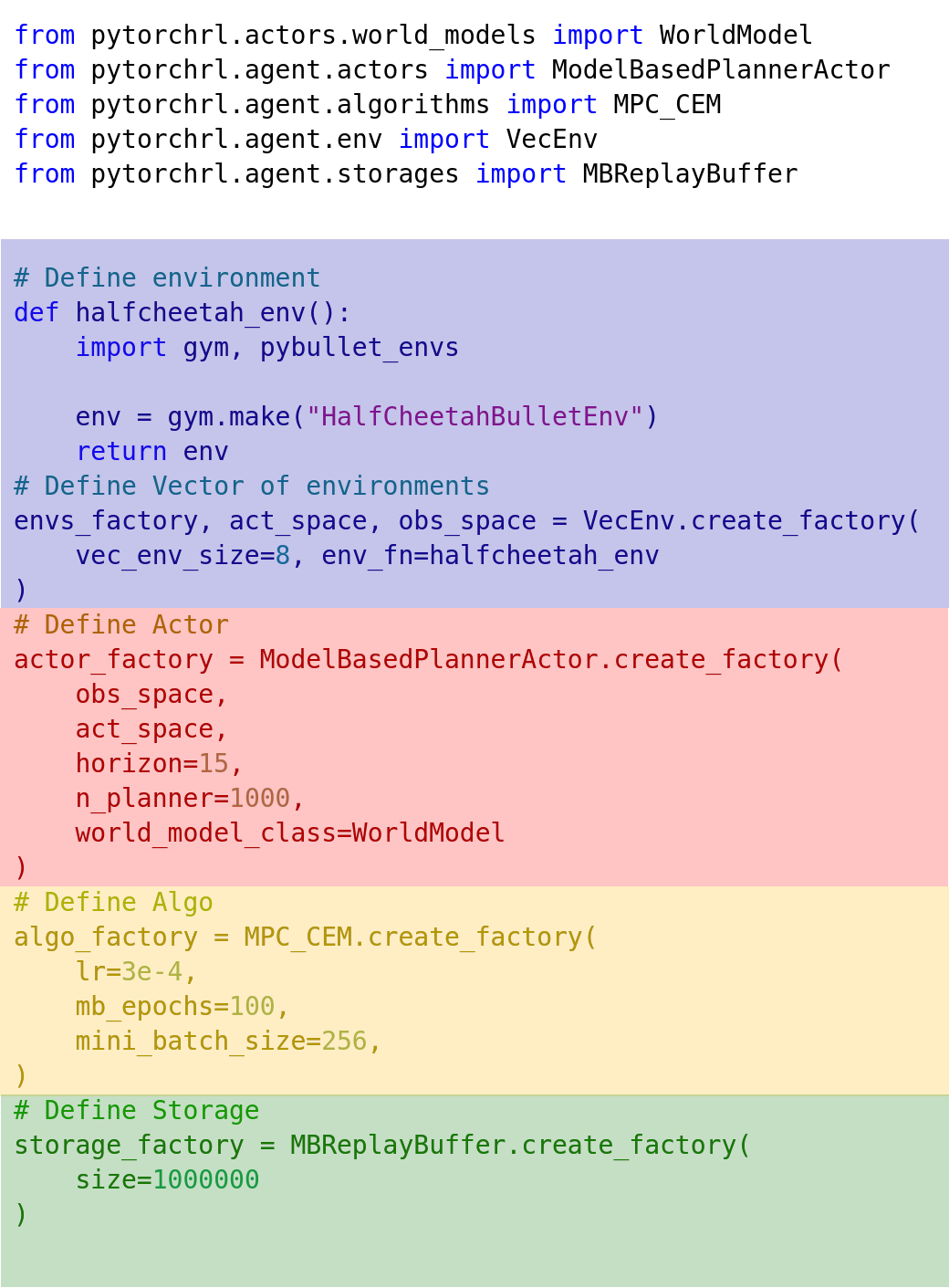}
  \caption{MPC example}\label{fig:mpc_example}
\endminipage
\end{figure*}

Among the libraries that offer modularity and distributed training, RLlib \cite{liang2017rllib} and RLgraph \cite{Schaarschmidt2019} utilize a distributed framework called Ray \cite{moritz2018ray}. This framework is characterized by logically centralized control, task parallelism, and resource encapsulation. In a similar vein, the ACME \cite{hoffman2020acme} library employs launchpad \cite{yang2021launchpad} for executing distributed training. Launchpad is a library specifically designed to simplify distributed programming by representing systems as graphs with nodes representing computational services. Nonetheless, there are two commonalities shared by all these approaches. Firstly, they tend to employ high-level implementations, instantiating agents through single function calls that accept a multitude of parameters and handle agent composition "under the hood". Secondly, these approaches abstract the configuration aspects of the distributed scheme from the user. Such design choices can ensure the compatibility of method components and simplify the nuances of distributed scaling. However, they also make it more difficult to understand which code is being executed, hinder code reusability, increase its complexity, and remove fine-grained control over the distributed scheme definition from the user, thus limiting the available options. The absence of these features complicates the tasks of many RL practitioners and researchers. Finally, TorchRL \cite{TorchRL} provides a limited selection of standalone composable distributed components tailored for specific use cases. We believe that our design approach could be employed to expand its range of distributed components, thus encompassing all possible RL use cases.

\section{Methods}

In this section, we introduce the interchangeable agent components that provide a high level of flexibility to define and extend RL agents. These components can be easily customized to suit the requirements of specific RL problems. Additionally, we delve into the rationale behind our distributed design choices and present a detailed overview of the specific distributed components. We also provide implementation details of these components, showcasing how the combination of agent-reusable components and scheme-reusable components empowers practitioners with a remarkable degree of flexibility in two crucial dimensions: agent definition and training scheme definition. 

\subsection{Agent Composability}

Our first design choice is to allow for agent composability. The main constituents of RL agents can be selected from a pool of options, and the resulting set of components can be seamlessly combined to construct a working RL agent. Each component can be selected independently of the rest and any specific component can be replaced by other candidates of the same type, resulting in a different agent but precluding execution errors. We distinguish between four types of agent components implemented as different Python classes, the combination of which creates a complete RL agent: 

\textbf{Actors.} These components implement the deep neural networks used as function approximators in the form of PyTorch nn.Modules and provide the methods to update them and generate predictions (most notably, but not exclusively, next action predictions). Currently, three classes are available for implementation: \texttt{OnPolicyActor}, \texttt{OffPolicyActor}  and \texttt{ModelBasedPlannerActor} for on-policy and off-policy and model-based RL methods, respectively.

\textbf{Storages.} These components handle the storage, processing, and retrieval of environment transition data. For on-policy algorithms, the storage options include \texttt{VanillaOnPolicyBuffer}, \texttt{GAEBuffer} (Generalized Advantage Estimation \cite{gae}), and \texttt{VTraceBuffer} \cite{espeholt2018impala}. For off-policy algorithms, the storage options encompass \texttt{ReplayBuffer}, \texttt{NStepBuffer} (for multi-step TD-learning), \texttt{PERBuffer} (Prioritized Experience Replay \cite{schaul2016prioritized}), \texttt{HERBuffer} (Hindsight Experience Replay \cite{andrychowicz2018hindsight}), \texttt{EREBuffer} (Emphasizing Recent Experience \cite{wang2019boosting}). Finally, for model-based algorithms \texttt{MBReplayBuffer} (Model-Based RL Replay Buffer) is available.

\textbf{Algorithms.} These components manage loss and gradient computations from data. The Algorithms include the on-policy methods \texttt{A2C} (Advantage Actor-Critic 
 \cite{mnih2016asynchronous}) and \texttt{PPO} (Proximal Policy Optimization \cite{schulman2017proximal}), for off-policy methods \texttt{DQN} \cite{mnih2015human} (incorporating improved versions like Rainbow \cite{hessel2017rainbow}), \texttt{DDPG} (Deep Deterministic policy gradient \cite{ddpg}), \texttt{TD3} (Twin Delayed Deep Deterministic policy gradient \cite{td3}), \texttt{SAC} (Soft Actor-Critic \cite{Haarnoja2018SoftAA}) ,and \texttt{MPO} (Maximum a Posteriori Policy Optimisation \cite{mpo}) 

\textbf{Envs.} Vectorized environments to make more efficient use of computing resources during inference time. Can work with any Gym-based environment. Currently, the environments accessible for training include OpenAI \texttt{gym} \cite{gym} (including MuJoCo \cite{mujoco} and Pybullet \cite{coumans2017pybullet}), Obstacle Tower \cite{juliani2019obstacle}, Animal Olympics, Habitat \cite{crosby2019animal} and Trifinger Real Robot \cite{wuthrich2020trifinger}.

Following a modular paradigm, new components can be created and combined with already existing ones as they are rooted in Python abstract superclasses defining all methods requiring implementation by any subclass to ensure correct functionality. While our primary goal is not to function as a reference library, we offer component templates accompanied by comprehensive code documentation to simplify the process of creating new implementations.

Finally, in distributed training scenarios, it is frequently required to instantiate multiple instances of RL components in separate processes. For example, both data collection and gradient computation tasks necessitate a copy of the actor component. To address this challenge, we employ a straightforward solution. All agent components are equipped with a method called \emph{create\_factory()} that returns a function capable of instantiating a new copy of the component each time it is called. By utilizing these methods and passing them to the distributed components, we ensure a straightforward and modular bottom-up scheme definition while accommodating the need for interchangeable component instances. Code examples of the components for PPO, SAC, and Model-Based MPC can be found in figures \ref{fig:ppo_example}, \ref{fig:sac_example} and \ref{fig:mpc_example}, with color coding to depict different components.

\subsection{Distributed Scheme Composability}

In this subsection we detail the unique characteristics and challenges associated with distributed reinforcement learning (RL), to then leverage these insights to define our scheme components.

\subsubsection{Distributed Design Choices}

Deep RL algorithms are generally based on the repeated execution of three sequentially ordered operations: rollout collection, gradient computation, and policy update. In single-threaded implementations, all operations are executed within the same process, and training speed is limited by the performance that the slowest operation can achieve with the resources available on a single machine. Furthermore, these algorithms do not have regular computation patterns, i.e. while rollout collection is generally limited by CPU capacity, gradient computation is often GPU-bounded, causing inefficient use of the available resources.

To alleviate computational bottlenecks, a more convenient programming model consists of breaking down training into multiple independent and concurrent computation units called compute actors or simply actors, with access to separate resources and coordinated from a higher-level script. To prevent any confusion with the \emph{Actor} component in our RL agents, we refer to these compute units as Workers in the remainder of the paper. Even within the computational budget of a single machine, this solution enables more efficient use of computing resources at the cost of slightly asynchronous operations. Furthermore, if Workers can communicate across a distributed cluster, this approach allows to leverage the combined computational power of multiple machines. 

A Worker-based software solution offers two main design possibilities that define implementable training schemes:
 1) Any operation can be parallelized, executing it simultaneously across multiple Worker replicas. 2) Coordination between consecutive operations can be synchronous or asynchronous. In other words, it is possible to decouple two consecutive operations by running them in different Workers that never remain idle and coordinate asynchronously, achieving higher resource efficiency. Thus, specifying which operations are parallelized and which operations are asynchronous defines the training schemes that we can implement. Note that single-threaded implementations are nothing but a particular case in which any operation is parallelized or executed asynchronously. Breaking down RL training also enables a composable implementation, allowing for the flexible combination of different task-based components.

\begin{figure*}[ht!]
\minipage{0.45\textwidth}
    \includegraphics[width=8.5cm, height=7cm, frame]{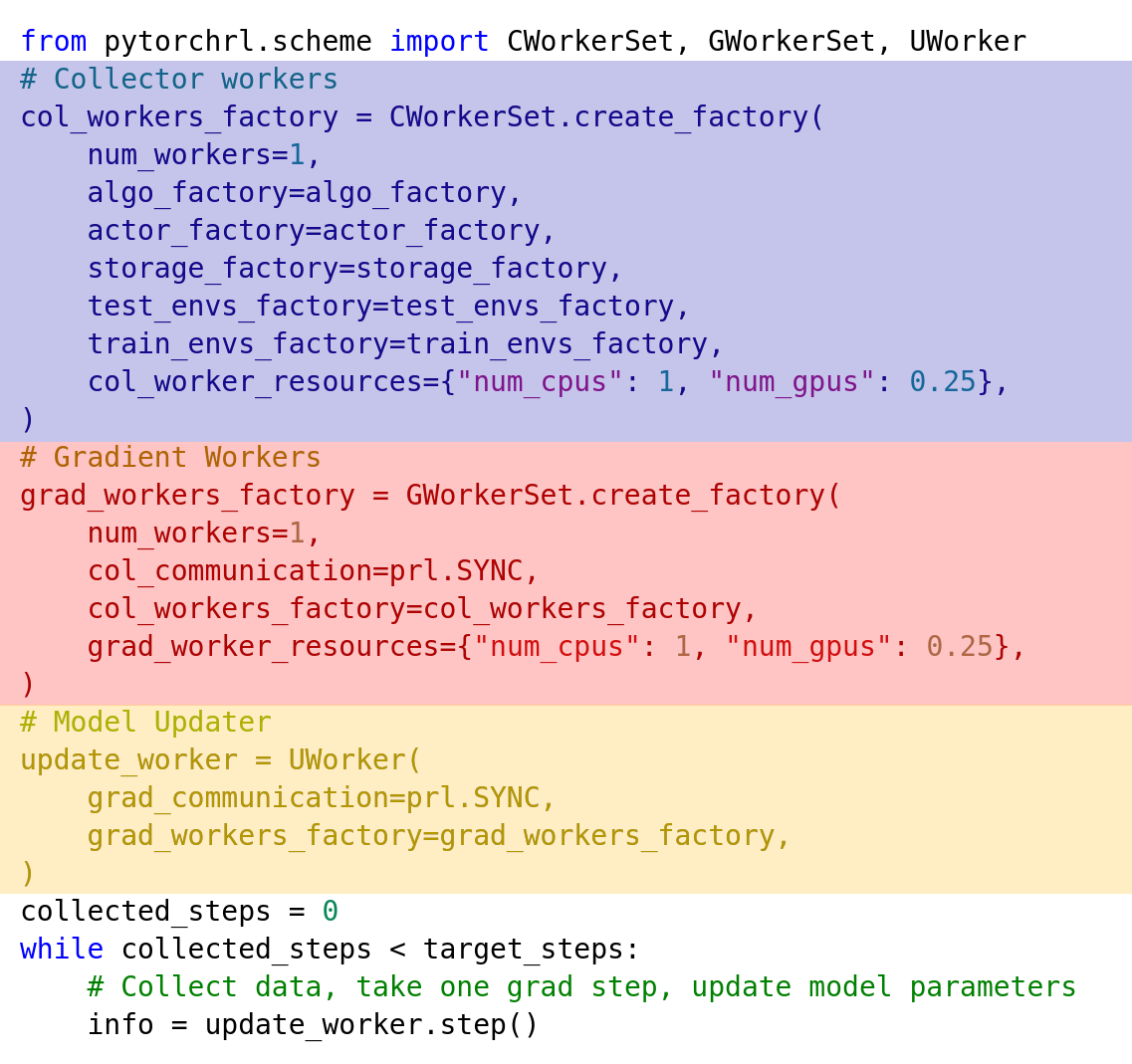}
  \caption{Scheme components offer a simple and composable way to define the number of workers of each type, worker resources, and worker coordination patterns with independence of agent components.}
 \label{fig:scheme_definition}
\endminipage\hfill
\minipage{0.48\textwidth}
  \centering
  \includegraphics[width=0.8\linewidth]{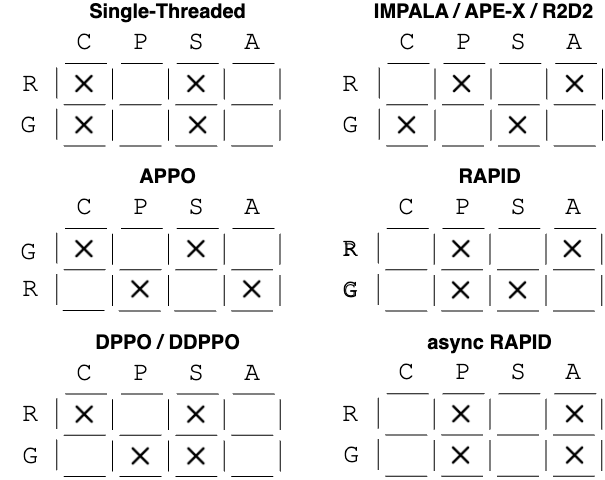}
  \caption{Execution of Rollouts collection (R) and Gradient computation (G) tasks can be either Centralised (C, executed only in 1 Worker) or Parallelized (P, multiple workers), and either Synchronous (S) or Asynchronous (A). Different parameter combination were used by DPPO \cite{heess2017emergence}, DDPPO \cite{wijmans2019dd}, APPO \cite{stooke2018accelerated}, IMPALA \cite{espeholt2018impala}, Ape-X \cite{horgan2018distributed} R2D2 \cite{kapturowski2018recurrent} or Rapid \cite{OpenAI_dota}.}
  \label{fig:distributed_schemes}
  \label{fig:scheme_options}
\endminipage
\end{figure*}

\subsubsection{Distributed Components}

The cornerstone of our distributed design is the Worker class, which is exclusively dedicated to a single operation among the three essential tasks: data collection, gradient computation, or model updates. Each variant of the Worker class is specifically tailored to handle a distinct operation, ensuring a clear and focused role for each Worker within the system. Based on the specific operation they perform, we categorize them as CollectionWorker, GradientWorker, and UpdateWorker, respectively. These Workers can be organized into sets to facilitate the parallel execution of tasks. As a result, the three key distributed components that form the foundation of any distributed architecture are the following: 

\textbf{Collector Workers Sets.} The collection worker plays a crucial role in gathering and transmitting data samples. In order to accomplish this task, it necessitates a copy of the environment, the actor, and the buffer. By consolidating multiple collection workers into a unified class, the collection worker effectively manages the process of data collection.

\textbf{Gradient Workers Sets.} A gradient worker receives data rollouts from a set of collection workers that it manages and computes gradients. It requires a copy of the storage, the actor and the algorithm components and provides the flexibility to specify whether the coordination between the collection workers is synchronous or asynchronous. In the synchronous case, the gradient worker waits for all the collection workers to complete their respective data collection before computing the gradients. Subsequently, an updated version of the actor is sent to all collection workers simultaneously. In the asynchronous case, the gradient worker utilizes the data as it becomes available from the collection workers. This ensures that the collection agents are continuously active, sending data and receiving slightly delayed versions of the actor, which we refer to as policy lag. This asynchronous approach enables efficient utilization of the collection workers without unnecessary idle periods. Multiple gradient workers are also grouped into a gradient worker set.

\textbf{Update Workers.} Finally, an update worker manages a set containing one or more gradient workers and solely requires a copy of the actor component to perform weight updates. It allows us to define the coordination strategy between the gradient workers. In the case of synchronous coordination, gradients from all workers are averaged and collectively applied to the neural networks. This ensures a synchronized update across all workers, promoting consistent learning progress. On the other hand, in asynchronous coordination, the workers operate continuously and some delay between the weights used for gradient computation and the weights to which the gradients are applied can is to be expected. This phenomenon is called gradient asynchrony. In our approach, we have made a deliberate decision to limit the size of the updater worker sets to a single worker. This choice is motivated by our focus on simplicity, excluding multi-agent applications from our current considerations.

By leveraging these components, researchers can design and configure distributed RL systems that effectively share the workload among multiple workers, optimizing the overall efficiency and performance of the training process.
For instance, as shown in Figure \ref{fig:scheme_options}, by defining a collection worker set consisting of multiple workers and configuring them to communicate asynchronously, one can achieve schemes such as IMPALA, APE-X, or R2D2. Alternatively, employing multiple gradient collectors with synchronous coordination allows for approaches like DPPO and DDPPO to be utilized.
We have chosen to implement our Workers utilizing the Ray \cite{moritz2018ray} distributed framework, which provides a robust and efficient infrastructure for distributed computing. Please note that the modularity of this approach would allow scheme components to be replaced by others. For example, alternative components utilizing backends other than Ray, such as Submittit \cite{submitit}, could be incorporated into the pool of available options to collect, compute gradients, or update networks. 

\begin{figure*}
    \centering
    \centerline{\includegraphics[scale=0.20]{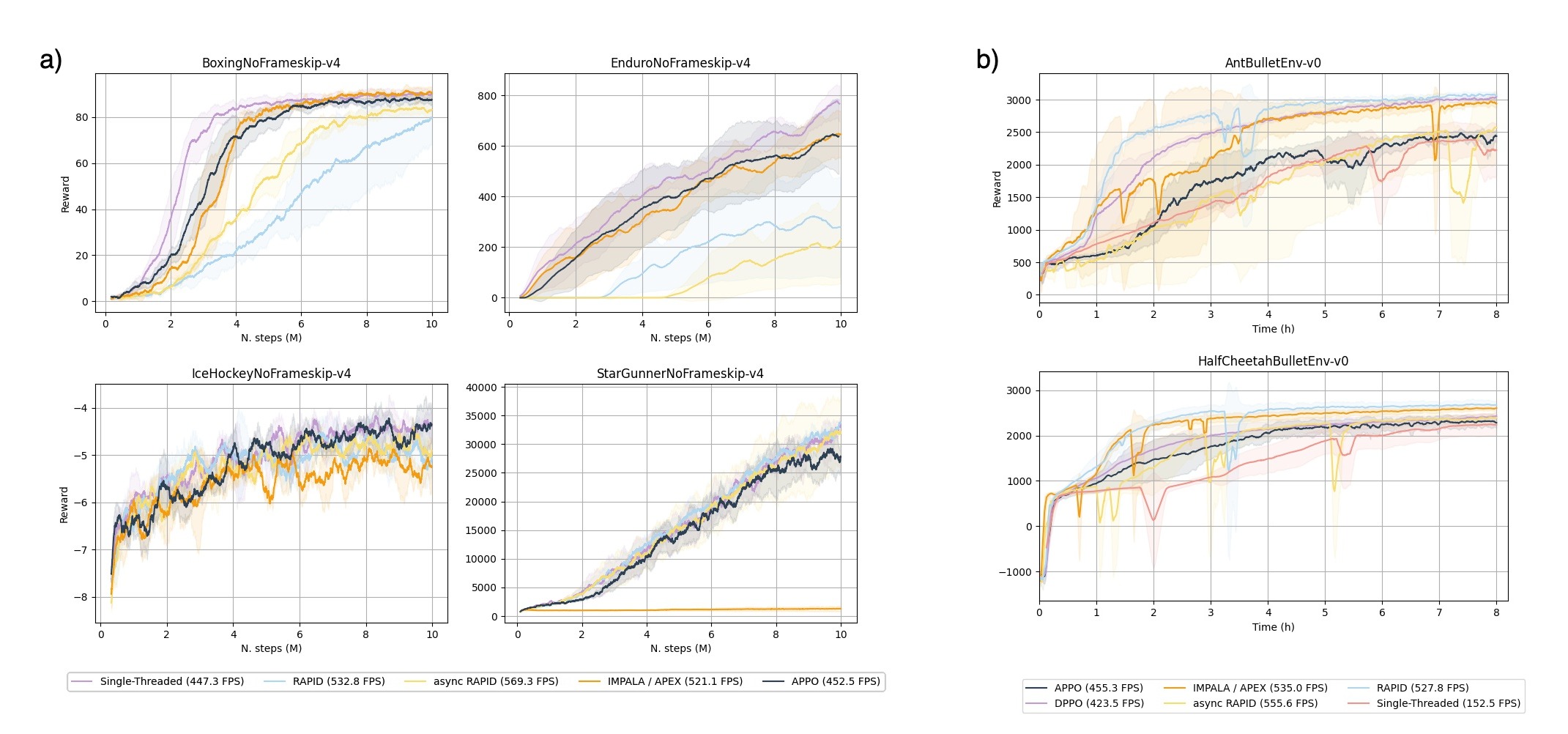}}
    \caption{\textbf{Experimental results on Atari 2600 with PPO-based agents (a) and Pybullet with SAC-based agents (b) environments with different distributed training schemes.} We train each scheme 3 times, with different random seeds and with random weight initialization, and plot the mean training curve. Plots were smoothed using a moving average with a window size of 100 points.}
    \label{fig.atari}
    \vspace{-4mm}
\end{figure*}

In summary, we formulate a comprehensive definition of a training process using a bottom-up approach. Initially, we select the agent components. Multiple example are shown in figures \ref{fig:ppo_example}, \ref{fig:sac_example} and \ref{fig:mpc_example}. Next, as illustrated in figure \ref{fig:scheme_definition} we establish the collector workers, followed by the gradient workers, and finally the update worker. Different scheme architectures can be configured as explained in Figure \ref{fig:scheme_options}. It is important to note that the composition of workers follows a hierarchical structure. Specifically, a gradient worker can coordinate multiple collection workers, and an update worker can manage multiple gradient workers.

\section{Experiments}

Distributed training schemes offer the advantage of significantly reducing training times, making them an appealing choice for accelerating the learning process. However, it is important to recognize that these schemes come with their own set of challenges and deviations from the original problem formulation. One common consequence is the introduction of policy lag when data collection is asynchronous, where there is a delay between the policy version used for data collection and the one used for gradient computations. Another deviation is gradient asynchrony, where the delay occurs between the policy version used for gradient computation and the application of the computed gradients. Additionally, certain schemes may alter the effective value of the batch size, further modifying the training process. These deviations pose a challenge in selecting the most suitable training architecture for each experiment.

Access to a variety of training scheme options, including both distributed and non-distributed approaches, allows practitioners to explore to identify the most appropriate scheme for each reinforcement learning application. In this section, we test whether or not our design choices allow for flexible experimentation at different levels of execution with independence to the agent components.

\subsection{Single Machine Benchmarks}

We validate our design choices by training a number of PPO-based agents on Atari 2600 games from the Arcade Learning Environment \cite{bellemare2013arcade}. We combine the agent with several architecture configurations including a Single-Threaded scheme and other schemes presented in Figure \ref{fig:scheme_options}. We refer to each scheme by the name of the original work that introduced it. We keep the number of collection and gradient Workers to 1, with the idea to test if training speed arises solely from decoupling operations. We measure training speed as the number of environment frames being processed, on average, in one second of clock time, or Frames-Per-Second (FPS). 
We fix all agent hyperparameters to be equal to the experiments presented in \cite{schulman2017proximal}, the original work presenting PPO, and use the same policy network topologies. We process the environment observations following \cite{mnih2015human} and train each architecture three times on each environment under consideration with randomly initialized policy weights every time. In \cite{schulman2017proximal} the authors used vectorized environments of size 8 to collect multiple rollouts in parallel. We do the same and use a different seed for each environment. Results are presented in Figure \ref{fig.atari}a.
Our single-threaded agents, and most of our other agents, present training curves close to those from \cite{schulman2017proximal}. Notably, even within the exact same computational budget, by moving from a completely sequential execution of operations (single-threaded) to a completely asynchronous one (async Rapid), we achieve a speed increase of over 25\%, from 447,34 FPS to 569,29 FPS. 

We also validate the flexibility of our approach in several experiments on a number of PyBullet \cite{coumans2017pybullet} environments with a SAC-based agent. It is common for RL agents to be bounded by single GPU memory, even when the machine has RAM memory and CPU resources available. However, employing a worker-based approach enables the allocation of distinct workers to different GPUs. Therefore, we design an experiment to test speed increases in a single machine using multiple GPUs and experiment with schemes that have $2$ gradient Workers managing $2$ data collection workers each.
We train each architecture $3$ times in each environment under consideration for 8h, using $3$ different random environment seeds and random parameter initialization in each case. Results are displayed in Figure \ref{fig.atari}b.
Hyperparameters can be found in the appendix material. 
We observe faster convergence to similar or superior reward values compared to the single-threaded experiment in almost all the cases. As the sole exception, architectures that introduce gradient asynchrony do not seem to improve the single-threaded training for the AntBulletEnv-v0, yet they do so for the HalfCheetahBulletEnv-v0 environment. Additionally, all architectures achieve training speed increases between $2.77$ and $3.64$ times that of the single-threaded baseline.

\subsection{Multiple Machines Benchmarks}

The rest of our experiments were conducted in the Obstacle Tower Unity3D challenge environment \cite{juliani2019obstacle} in order to test on more challenging environments. Obstacle Tower is a procedurally generated 3D world where an agent must learn to navigate an increasingly difficult set of floors (up to 100) with varying lighting conditions and textures. Each floor can contain multiple rooms, and each room can contain puzzles, obstacles, or keys enabling one to unlock doors. The number and arrangement of rooms on each floor for a given episode can be fixed with a seed parameter. This environment is far more complex than the previously tested and requires significantly more data only to start solving the initial rooms.

%
\begin{figure*}[th!]
    \centerline{\includegraphics[scale=0.30]{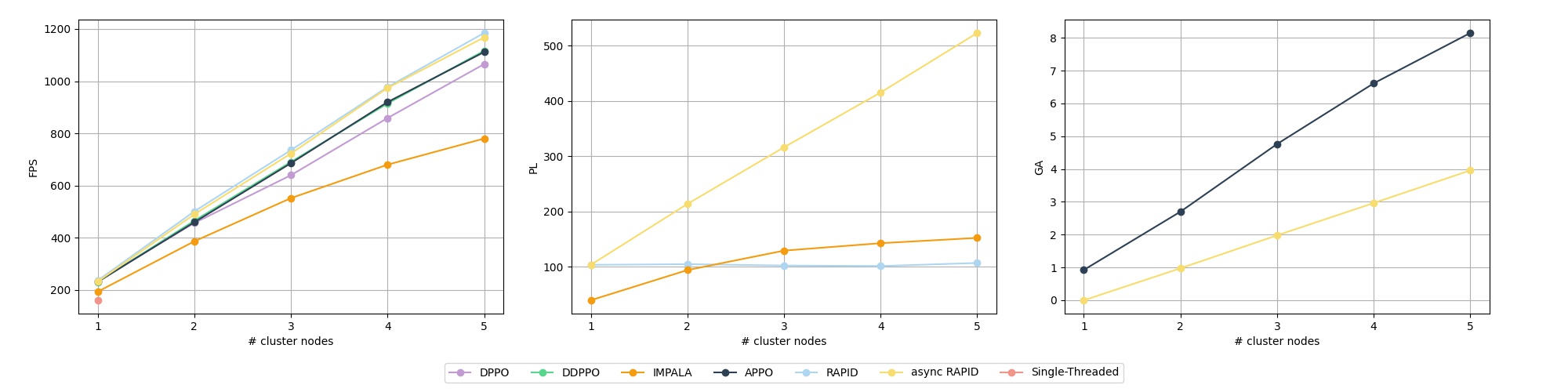}}
    \caption{\textbf{Scaling Performance with multiple machines.} Frames-per-Second (FPS), Policy Lag (PL) and Gradient Asynchrony (GA) of different training schemes in increasingly large clusters. We do not plot curves that remain constant to 0 as cluster size increases.}
    \label{fig.scale_plot}
    \vspace{-3mm}
\end{figure*}

\begin{figure*}[h!]
    \centerline{\includegraphics[scale=0.18]{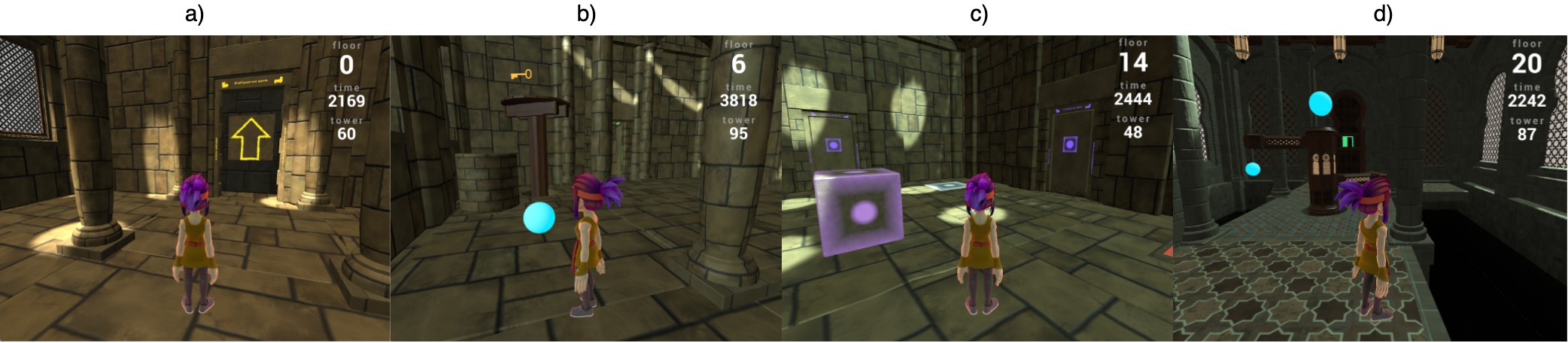}}
    \caption{\textbf{Obstacle tower Unity3D challenge environment.}}
    \label{fig.obstacle}
\end{figure*}

We design our second experiment to test the capacity of our implementations to accelerate training processes in increasingly large clusters. We benchmark the FPS, the average policy lag (PL) and the average gradient asynchrony (GA) when training a PPO-based on-policy agent in clusters composed of $1$ to $5$ machines. We experiment with the same training scheme as in the previous section but also include a variant of DPPO, DDPPO. In DDPPO, once a predefined percentage of the parallel data collection Workers have finished their task, the straggling Workers are forced to early stop their task and the training proceeds. We set a preemption threshold of $80\%$, a parameter available in our gradient workers. Agent hyperparameters are held constant throughout the experiment. For each specific cluster size and training scheme combination, we select the number of collection and gradient Workers that maximize the training speed. We used machines with 32 CPUs and 3 GPUs, model GeForce RTX 2080 Ti. We could use two GPUs to obtain similar results if the environment instances could be executed in an arbitrarily specified device. However, currently, Obstacle Tower Unity3D challenge instances run by default in the primary GPU device, and thus we decide to devote it exclusively to this task. Our results are plotted in Figure \ref{fig.scale_plot}.

One noteworthy aspect is that we conducted this experiment with minimal adjustments to our training script, highlighting the advantages of modular code. By keeping the agent components intact and solely modifying the configuration of our scheme components, we achieved great flexibility and control while effectively managing code complexity. This capacity to seamlessly test multiple distributed schemes offers significant benefits in understanding the behaviors exhibited by different training approaches, empowering practitioners to make informed design decisions. 

\subsection{Scaling to solve computationally demanding environments}  

\begin{figure*}[ht!]
    \centering
    \centerline{\includegraphics[scale=0.3]{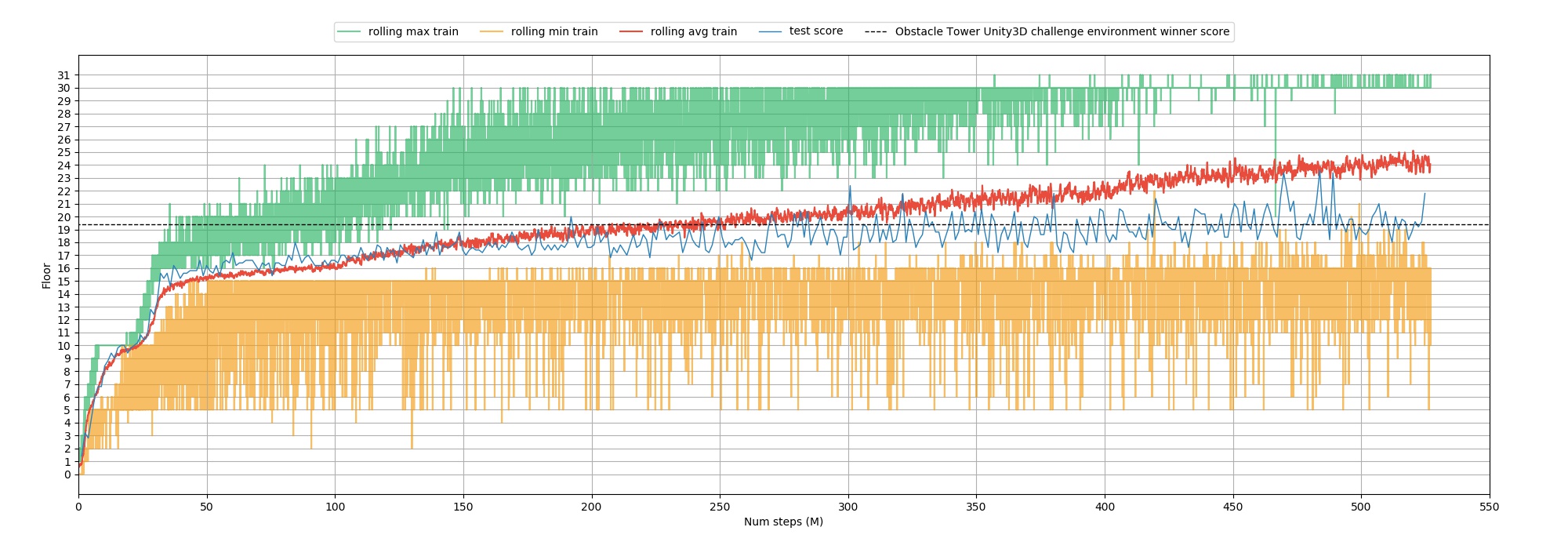}}
    \caption{\textbf{Train and test curves on the Obstacle tower Unity3D challenge environment.} We use a rolling window size of size 20 to plot the maximum and the minimum obtained scores during training, and a window of size 250 for the training mean. A video recording of the best policy  is available at \cite{obstacle_tower_demo}}
    \label{fig.obstacle_results}
    \vspace{-5mm}
\end{figure*}

In our last experiment, we aim to demonstrate a training scenario that requires the integration of all key features sought by practitioners in RL libraries. These features include the ability to flexibly experiment with agent components, experiment with scheme components, and scale the solution without the need for reimplementation. 
The obstacle environment consists of navigating a series of rooms connected by one or more doors. The action space is discrete (e.g. move forward, turn right, jump, move back, etc). The observation is an image of 84x84x3, no map is available. From floors 1-5 the only task is to be able to navigate rooms towards the next floor (see Figure \ref{fig.obstacle}a). There is a time limit that is extended every time a floor is completed. When a floor is passed the environment returns a +1 reward, and intermediate doors are given a smaller reward. There can be closed rooms where the player needs to navigate back to the previous room and pass another door. From floors 5 to 10, some doors are closed and require a key that can be picked up somewhere (see Figure \ref{fig.obstacle}b). Sometimes the key is on the ground but sometimes picking the key requires  coordinated movements to jump up moving platforms or stairs.
From floors 10-15 a box needs to be pushed over a platform on the ground in order to open doors (see Figure \ref{fig.obstacle}c). Both the position of the box and platform are random and can be in any room on the floor, so the player needs to navigate there first. This level was only solved by two participants in the original challenge. Above 15, the player can also fall into holes and die, the texture of the walls can change and  rooms can become very complex (see Figure \ref{fig.obstacle}d). 

We experiment with diverse agents and finally settle for an on-policy agent composed of our \emph{PPO}, \emph{GAEBuffer} and \emph{OnPolicyActor} components, and use \emph{VecEnv} to define environment vectors of size $16$. 
As a training scheme, we use a distributed RAPID-like scheme as it offers high training speed while keeping PL and GA metrics low and stable. Nonetheless, we observe low sensitivity to the increase of these metrics in The Obstacle Tower Unity3D environment. We use $4$ gradient Workers and $8$ collection Workers. Specific hyperparameter values, neural network typologies and a description of the reward shape mechanism are provided in the supplementary material.
We train an RL agent on the Obstacle Tower Unity3D environment over a cluster with 64 CPU cores and 6 GPUs and compare it against the state-of-the-art, obtained during the challenge competition organized by Unity Technologies upon the environment release. We train an agent on a fixed set of seeds [0, 100) for approximately $11$ days and test its behavior on seeds $1001$ to $1005$, a procedure designed by the authors of the challenge to evaluate weak generalization capacities of RL agents \cite{juliani2019obstacle}. The maximum average test score is $23.6$, which supposes a significant improvement with respect to $19.4$, the previous state-of-the-art obtained by the winner of the competition. Our final results are presented in Figure \ref{fig.obstacle_results} showing that we are also consistently above $19.4$. A video recording of our model performance is available at \cite{obstacle_tower_demo}, showing that the agent reached high levels of skill after being trained with over 500 million frames of experience.





\section{Conclusion}

In this paper, we present an approach that enables the flexible and explicit definition of RL agents and training schemes across one or more machines. By introducing modularity, code reusability, and composition into the domain of distributed training, we address a longstanding challenge wherein such aspects have traditionally been handled internally and abstracted away from users in popular RL libraries. Through our experimental results, we showcase the efficacy of our design choices in granting users complete freedom to explore distributed schemes while maintaining fine-grained control. Moreover, we demonstrate the practical application of our approach by significantly improving the state-of-the-art performance in a challenging RL problem. We firmly believe that the integration of scalable training into highly modular libraries is a crucial step forward. We posit that our work offers valuable insights and practical guidance to those interested in contributing to the development of the next generation of libraries that cater to the needs of a broader user base. Our codebase is available on \href{https://anonymous.4open.science/r/pytorchrl-8E00}{github}.

\bibliography{journal} 

\clearpage


%

\appendices
\section{Experiments on Atari 2600 environments}

We used the ensuing parameters, following \cite{schulman2017proximal}. We also used the same CNN network architecture:

\begin{table}[h!]
\caption{PPO + GAE Agent - hyperparameters values.}
\label{sample-table}
\vskip 0.15in
\begin{center}
\begin{small}
\begin{sc}
\begin{tabular}{l|c}
\toprule
  Parameter & Value \\ \midrule
  Adam lr & $2.5e-4$ \\
  clip-param & 0.15 \\
  gamma & 0.99 \\ 
  frame-skip & 4 \\
  frame-stack & 4 \\
  num-steps & 128 \\
  num-mini-batch & 4 \\
  num-epochs & 3 \\
  entropy-coef & 0.01 \\
  value-loss-coef & 1.0 \\
  environment vector length & 8 \\
  gae-lambda & 0.95 \\
  max-grad-norm & 0.5 \\
\bottomrule
\end{tabular}
\end{sc}
\end{small}
\end{center}
\vskip -0.1in
\end{table}

The experiment was executed with the following training script. During training, the learning rate and the PPO clipping parameter are decayed also following \cite{schulman2017proximal}.

\section{Experiments on PyBullet environments}

We used the following parameters, hand-picked to perform well on the single-threaded regime. As a feature extractor, we used a multilayer perceptron with two hidden layers of 256 each neurons and ReLU activation function:

\begin{table}[h!]
\caption{SAC Agent - hyperparameters values.}
\label{sample-table}
\vskip 0.15in
\begin{center}
\begin{small}
\begin{sc}
\begin{tabular}{l|c}
\toprule
  Parameter & Value \\ \midrule
  Adam lr Q networks & $1e-3$ \\
  Adam lr policy & $1e-4$ \\
  Adam lr alpha & $1e-05$ \\
  gamma & 0.98 \\
  initial alpha & 0.2 \\
  environment vector length & 1 \\
  polyak update  & 0.995 \\
  buffer size & 1500000 \\
  mini-batch size & 256 \\
  start steps & 10000 \\
  num updates & 32 \\
  updateevery & 128 \\
\bottomrule
\end{tabular}
\end{sc}
\end{small}
\end{center}
\vskip -0.1in
\end{table}

\section{Obstacle Tower 3D Hyperparameters and Rewad Shaping}

For the Obstacle Tower 3D environment we use the network architecture proposed in \cite{espeholt2018impala} but we initialize its weights according to Fixup \cite{zhang2019fixup} and end our network with a gated recurrent unit (GRU) \cite{gru} with a hidden layer of size 256 neurons. Gradients are computed using Adam optimizer \cite{kingma2014adam}, with a starting learning rate of 4e-4 decayed by a factor of 0.25 both after 100 million steps and 400 million steps. The value coefficient, the entropy coefficient, and the clipping parameters of the PPO algorithm are set to 0.2, 0.01, and 0.15 respectively. We use a discount factor gamma of 0.99. Furthermore, the horizon parameter is set to 800 steps and rollout collections are parallelized using environment vectors of size 16. Gradients are computed in mini-batches of size 1600 for 2 epochs. Finally, we use generalized advantage estimation \cite{gae} lambda of 0.95, frame skip $2$ and frame stack $4$. We restart each new episode at a randomly selected floor between $0$ and the higher floor reached in the previous episode. During training, we restart each new episode at a randomly selected floor between $0$ and the higher floor reached in the previous episode and regularly save policy checkpoints to evaluate the progression of test performance. Test performance is measured as the highest averaged score on the five test seeds obtained after $5$ attempts, due to some intrinsic randomness in the environment. We also experiment with several reward-shaping modifications, which we detail in the appendix.

We experiment with several reward-shaping modifications and finally find a surprisingly simple reward structure that helps our agent reach higher floors. The extrinsic reward upon the agent completing a floor is $+1$, and $+0.1$ is provided for opening doors, solving puzzles, or picking up keys. We additionally reward the agent with intrinsic reward: an extra $+1$ to pick up keys, $+0.002$ if a box is in its field of view, $+0.001$ if a platform is in its field of view, $+1.5$ to place the boxes on the platform. We use a simple color detection on the observation image to classify if the box or the platform are in view. We also reduce the action set from the initial 54 actions to 6 (rotate the camera in both directions, go forward, and go forward while turning left, turning right or jumping). The code of our training script is provided in the supplementary material. This simple set of modifications is capable to reach the state-of-the-art by brute force sampling of the environment for hundreds of millions of steps.

\ifCLASSOPTIONcaptionsoff
  \newpage
\fi

\end{document}